\documentclass[conference]{IEEEtran}
\IEEEoverridecommandlockouts
\usepackage{cite}
\usepackage{amsmath,amssymb,amsfonts}
\usepackage{algorithmic}
\usepackage{graphicx}
\usepackage{textcomp}
\usepackage{acronym}  
\acrodef{ANN}{Artificial Neural Network}
\acrodef{STDP}{Spike-Timing-Dependent Plasticity}
\acrodef{PSP}{Postsynaptic Potential}
\acrodef{SRM}[SRM\textsubscript{0}]{Spike Response Model}
\acrodef{SNN}{Spiking Neural Network}
\acrodef{LIF}{Leaky Integrate-and-Fire}
\acrodef{vRD}{van Rossum Distance}
\acrodef{FILT}{FILTered-error}
\acrodef{DLS}{Digital Learning System}
\usepackage[separate-uncertainty=true]{siunitx}  
\usepackage[colorinlistoftodos]{todonotes}

\def\BibTeX{{\rm B\kern-.05em{\sc i\kern-.025em b}\kern-.08em
    T\kern-.1667em\lower.7ex\hbox{E}\kern-.125emX}}
\begin{document}

\title{Supervised Learning in Temporally-Coded Spiking Neural Networks with Approximate Backpropagation
\thanks{This research was supported in part by Seagate Technology, PLC. This research has also received funding from the European Union's Horizon 2020 Framework Programme for Research and Innovation under the Specific Grant Agreement No. 785907 (Human Brain Project SGA2).}
}

\author{Andrew Stephan, Brian Gardner, Steven J. Koester, \textit{Fellow, IEEE,} and Andr{\'e} Gr{\"u}ning
\thanks{Manuscript submitted \today.}
\thanks{The authors acknowledge the Minnesota Supercomputing Institute (MSI) at the University of Minnesota for providing resources that contributed to the research results reported within this paper. URL: http://www.msi.umn.edu}
\thanks{A. W. Stephan is with the Department of Electrical and Computer Engineering, University of Minnesota, Minneapolis, MN 55455 USA (e-mail:steph506@umn.edu).}
\thanks{B. Gardner is with the Department of Computer Science, University of Surrey, Guildford, UK (e-mail:b.gardner@surrey.ac.uk).}
\thanks{S. J. Koester is with the Department of Electrical and Computer Engineering, University of Minnesota, Minneapolis, MN 55455 USA (e-mail:skoester@umn.edu).}
\thanks{A. Gr{\"u}ning is with the Department of Electrical Engineering and Computer Science, University of Applied Sciences, Stralsund, Germany (e-mail:andre.gruening@hochschule-stralsund.de).}
}

\maketitle

\begin{abstract}
In this work we propose a new supervised learning method for temporally-encoded multilayer spiking networks to perform classification. The method employs a reinforcement signal that mimics backpropagation but is far less computationally intensive. The weight update calculation at each layer requires only local data apart from this signal. We also employ a rule capable of producing specific output spike trains; by setting the target spike time equal to the actual spike time with a slight negative offset for key high-value neurons the actual spike time becomes as early as possible. In simulated MNIST handwritten digit classification, two-layer networks trained with this rule matched the performance of a comparable backpropagation based non-spiking network.
\end{abstract}


\section{Introduction}
\label{sec:Introduction}
Backpropagation is a commonly used optimization procedure for \acp{ANN} in data classification, and has demonstrated particular success in recent years as increased processing speed has become available through massively parallel computing architectures. Despite this, the energy requirements of computing as applied to traditional \acp{ANN} continues to rise with the complexity of input data, limiting the versatility of machine learning in real-world applications. \acp{SNN}, which incorporate the actual firing times, or spikes, of simulated neurons as part of their operational model, have been identified as a means to provide more energy efficient, event-based computing, while still theoretically still being capable of greatly increased computational power over \acp{ANN} \cite{Maass1997}. Despite this, applying backpropagation to \acp{SNN} has proven to be challenging, since the spikes emitted by neurons have no smooth functional dependence on network parameters, making reliable gradient estimation an issue. Over the last decade there have been numerous attempts to address this spike-gradient challenge, using various approximations to producing gradients that can be differentiated with respect to network parameters.

One of the first examples in combining backpropagation with an \ac{SNN} is described in \cite{Bohte2002}, where a feedforward network containing a single hidden layer was trained to associate spatiotemporal input patterns with desired network responses. The network's error-function was defined in terms of the squared difference between desired and actual firing times in the output layer, upon which gradient descent was performed to obtain weight gradients in each layer. To solve for the gradient of a discontinuous firing time, the authors assumed a linear dependence of this quantity on its activating input for a sufficiently small temporal region. While this approximation displayed success on limited, small-sized datasets, some drawbacks were noted such as the rule's reliance on small learning rates to ensure eventual convergence, as well as its limitation to just learning single-spike responses in each layer.

A later approach to combining backpropagation with \acp{SNN} has been proposed in \cite{Sporea2013} for learning desired input-output spike pattern associations. A first step taken by the authors of this study was to relate a previous spike-based supervised learning method, called ReSuMe \cite{Ponulak2010}, to an error-minimization procedure based on gradient descent. Secondly, the activations of neurons in a multilayer network were formulated in terms of their instantaneous firing rates, which were assumed to have a linear functional dependence on their previous layer inputs. Hence, by applying backpropagation to this approximately-equivalent rate-based network, and then substituting neuronal firing rates for their actual spike trains, ReSuMe could be applied to multilayer learning. As a proof-of-concept the authors demonstrated the success of this rule on a small, benchmark data classification task. The main limitation of this approach relates to the assumption that neurons in the network can be treated as linear activations, which potentially diminishes the ability of the network to solve linearly non-separable classification tasks.

A further example of backpropagation in \acp{SNN} was shown in \cite{Gardner2015}, which extended a maximum-likelihood learning scheme from single to multilayer network structures. Building on the work of \cite{Pfister2006}, this study considered networks containing stochastic spiking neurons, such that spike-gradients could be approximated by their expectations of generating sequences of output spikes. This probabilistic approach has the advantage of still retaining the non-linear behavior of the network, since the actual firing times of hidden neurons are taken into account when computing weight gradients. The capability of this rule was demonstrated quite extensively on a variety of associative learning tasks, although its technical performance on real-world datasets has not been tested.

A more recent examination of backpropagation in \acp{SNN} has taken advantage of the analytical tractability of non-leaky integrate-and-fire neurons constrained to single-spike responses \cite{Mostafa2017}; Specifically, the author arrived at a closed-form solution for backpropagation in a multi-layered network of such neurons, making it more easily extendable to \acp{SNN} containing more than one hidden layer. Furthermore, since just the first output spikes emitted by neurons were considered, the network was capable of classifying MNIST digits with minimal delay, with low test errors of around 3 \%. The main limitation of this study relates to its restriction to just single spikes per neuron in all layers.

One of the most recent approaches to formulating spike-based backpropagation has relied on the use of `surrogate gradients' in order to establish smooth functional derivatives with respect to network parameters \cite{Zenke2018}. Termed SuperSpike, this learning rule works to minimize a spike train (dis)similarity measure called the \ac{vRD} \cite{vanRossum2001} which is a function of the distance between target and actual output spike trains in the final layer of a multilayer SNN. In estimating the gradient of a spike train, the author used the slope of a sigmoid as an auxiliary function. In terms of the rule's performance, good accuracy was demonstrated in terms of its ability to precisely match arbitrary spatiotemporal spike patterns. As is the limitation with most spike-based backpropagation rules, SuperSpike relies on the presence of hidden spikes in order to allow the computation of weight gradients.

The above studies highlight the typical approach that is taken to training multilayer \acp{SNN}: selecting an appropriate output objective function, and establishing a smooth functional dependence with respect to the network's parameters such that backpropagation can be applied. Although this procedure has displayed success on certain learning tasks, there remains the issue that computing weight gradients in such a way is usually computationally expensive, and in most cases difficult to scale when applied to deeper network architectures. Certain model simplifications such as those described in the work of \cite{Mostafa2017} help mitigate these issues, although this then trades off the increased computational capability afforded by \acp{SNN} when compared against \acp{ANN}.

Generally-speaking, training \acp{SNN} using an unsupervised learning method such as \ac{STDP} \cite{Gerstner2002} or a single-layer, supervised learning rule \cite{Pfister2006,Ponulak2010,Florian2012,Mohemmed2012,Memmesheimer2014,Gardner2016} is inherently advantageous in terms of computational efficiency over spike-based backpropagation when computing weight gradients. This follows from their reliance on immediately available learning factors: specifically pre- and postsynaptic activity variables, whereas backpropagation also explicitly depends on an error signal that requires a series of backpropagation steps to be computed, starting from the final layer. Furthermore, as described previously, backpropagation applied to \acp{SNN} also relies extensively on approximate spike-gradients, which for the most part is avoided with localized learning methods. Despite this, it is clear that backpropagation is better suited to training multilayer networks as applied to data classification tasks. This is apparent when considering that individualised error signals are backpropagated through the network, informing each synapse of its specific contribution to the network's overall accuracy during training.
Hence, it becomes reasonable to suppose that a hybrid approach to \ac{SNN} training combining localized layer-wise learning with generalised feedback signalling is capable of network performance approaching that of backpropagation, but without the sacrifice in computational cost.



In this paper we propose a novel hybrid learning algorithm for multilayer \acp{SNN} that uses one learning process for all of its layers. The algorithm builds on the work of \cite{Gardner2015,Gardner2016}, beginning with a supervised learning method that circumvents the discontinuous spike-gradient issue by taking a maximum-likelihood approach to learning, which is then taken in the limit of a deterministic system to allow for more precise network responses \cite{Gardner2016}. With respect to each layer, the algorithm computes weight gradients by combining a set of localized learning factors with a simple reinforcement signal that carries a summary of the behavior of downstream layers in the network. This signal is condensed into a set of analog values associated with each postsynaptic neuron. This allows the learning algorithm to make decisions while armed with some knowledge of the global impact of each synapse without needing to backpropagate a complex error gradient function.

The rest of this paper is organized as follows. In Section \ref{sec:Network Model} we describe the mathematical model of the neurons and input encoding, as well as the basic network structures employed. The learning algorithm is explained in Section \ref{sec:Learning Theory}, and the simulation results given in Section \ref{sec:Simulation}. We conclude with a discussion in Section \ref{sec:Discussion}.

\section{Network Model}
\label{sec:Network Model}
\subsection{Spiking Neuron Model}
This work uses the simplified \acf{SRM} to describe the dynamics of simulated neurons \cite{Gerstner2002}. Specifically, a postsynaptic neuron's membrane potential $u_i$ at time $t$ is described by
\begin{gather}
u_i(t|\boldsymbol{x},y_i) = \sum_j w_{ij}\sum_{t_j \in x_j} \epsilon (t - t_j) + \sum_{t_i \in y_i} \kappa (t - t_i),
\end{gather}
where $x_j \in \mathbf{x}, x_j = \{t_{j,1}, t_{j,2}, \dots\}$, is the spatiotemporal spike pattern of all presynaptic neurons indexed by $j$, and $y_i = \{t_{i,1}, t_{i,2}, \dots\}$ is the sequence of emitted spikes, or spike train, of the postsynaptic neuron with index $i$. On the right-hand side of the equation, $w_{ij}$ is the synaptic weight between pre- and postsynaptic neurons $j$ and $i$, respectively, $\epsilon$ describes the form of the \ac{PSP} evoked at the postsynaptic neuron in response to a single presynaptic spike and $\kappa$ is the reset kernel. The first double sum determines the net weighted \ac{PSP} response due to all incoming spikes, while the second sum describes postsynaptic refractory effects due to the emission of output spikes. The $\epsilon$ kernel is given by
\begin{gather}
\epsilon(s) = \epsilon_0 \left[ \mathrm{exp} \left(-\frac{s}{\tau_m}\right) - \mathrm{exp} \left(-\frac{s}{\tau_s}\right) \right] \Theta(s),
\end{gather}
where $\tau_m = 10$ ms and $\tau_s = 5$ ms are the membrane and synaptic time constants, respectively, and the coefficient $\epsilon_0 = 4$ mV. $\Theta$ is the Heaviside step function.
The reset kernel is given by
\begin{gather} \label{eq:reset}
\kappa(s) = (u_r - V_t) \exp \left( -\frac{s}{\tau_m} \right) \Theta(s),
\end{gather}
where $u_r = 0$ mV and $V_t = 15$ mV are the reset and firing threshold potentials, respectively. In terms of spike-generation, the postsynaptic neuron fires a spike when its membrane potential exceeds $V_t$, immediately after which its potential is reset to $u_r$ in response to Eq~\eqref{eq:reset}.

\subsection{Network Structure}
We will employ two different network structures in the experiments carried out in this paper. The first consists of a single \ac{SRM} neuron with either one or two input channels, each of which contributes a single spike at a preset time. This layout will be used to study the firing responses of individual neurons subject to the training scheme that will be described in Section \ref{sec:Learning Theory}. The second network will be designed to test the performance of the proposed hybrid learning algorithm as applied to the MNIST handwritten digits classification problem, consisting of an input layer, at least one hidden layer and an output layer. The input layer fulfills the role of encoding digits as precisely-timed spikes, which shall be described in more detail in the following subsection. The hidden and output layers consist of \ac{SRM} neurons that perform computations on these inputs, which shall be trained using the rule proposed below. The hidden layers shall contain 100, 200 or 300 neurons, while the size of the output layer shall be held fixed at 10 neurons: each of which corresponds to one of the 10 handwritten digit classes. Whichever output neuron produces the first spike in response to a presented digit is used to decide the network's classification of the input. If no output neurons spike, no classification is made. The network is trained in 500 batches of 20 iterations each, as limited by available computing resources. Both the single-layer and multi-layer network layouts are simulated for 10 ms per iteration.

\subsection{Temporal Encoding of Input}
To solve the MNIST classification problem \cite{lecun1998} we must first translate the images into input spikes. Specifically, the input layer, or layer 0, is set up to contain 784 channels, corresponding to the 784 pixels in the 28x28 image with one-to-one association. Each channel is then driven to produce up to one spike, where the spike timing depends on its associated pixel intensity. The timings of spikes are determined as follows. First, pixels are normalized such that their values fall within the range [0, 1]. We then choose a Gaussian sensitivity function for each timing, inspired by the population encoding model in \cite{Bohte2002unsupervised}:
\begin{gather}
t^0_j(p_j) =
\begin{cases}
    T * \left[1 - \mathrm{exp} \left(-\frac{(p_j-1)^2}{2\cdot \sigma^2}\right) \right] & \text{for } p_j \ge p_t\\
    \infty &\text{for } p_j < p_t,
\end{cases}
\end{gather}
where $t^0_j(p_j)$ is the time of the input spike produced by channel $j$ in the $0$th layer as a function of the pixel strength $p_j$, $T = 10$ $ms$ is the duration of the simulated time and $\sigma = 0.5$ determines the breadth of the sensitivity function. Any pixel value below the threshold $p_t = 0.5$ is considered not to produce a spike, as represented by an infinite response time. For a pixel value greater than the threshold, as the value increases the spike timing occurs closer to $t^0_j(p_j) = 0$, while a decrease in its value delays the spike up to a maximum delay of almost \SI{4}{ms}. From a biological perspective, this rapid encoding scheme is supported by observations of neural populations in the brain being capable of encoding visual information using spikes occurring within a very small time window, on the order of around \SI{10}{ms} \cite{Hung2005}.

\section{Learning Theory}
\label{sec:Learning Theory}
\subsection{\acf{FILT} Learning Rule}
We start our analysis by first describing the \ac{FILT} synaptic plasticity rule, a supervised method which was originally derived in \cite{Gardner2016} to impose a target output spike train on an \ac{SRM} neuron. The distinction of this rule as compared with most other single-layer, spike-based learning rules is that, in the initial stage of its derivation, a stochastic firing rate substitution was used to circumvent the non-differentiable spike-gradient problem. Thereafter, in taking the limit of a deterministic spiking neuron with a fixed firing threshold, the following $\lambda$ learning window, as a function of the separation between target and actual output spike times, was derived:
\begin{gather}
\lambda(s) = 
\begin{cases}
  \epsilon_0 \left[ C_m \mathrm{exp} \left( -\frac{s}{\tau_m} \right) - C_s \mathrm{exp} \left( -\frac{s}{\tau_s} \right) \right] & \text{for } s > 0\\    
  \epsilon_0 (C_m - C_s) \mathrm{exp} \left( \frac{s}{\tau_q} \right)    & \text{for } s \le 0,
\end{cases}
\end{gather}
where $\tau_q = 10$ ms, $C_m = \frac{\tau_m}{\tau_m + \tau_q}$ and $C_s = \frac{\tau_s}{\tau_s + \tau_q}$. The $\tau_q$ term is a spike-linkage variable, describing the coincidence between target and actual output spike times. Hence, given an input spike pattern $\boldsymbol{x}$, and single target and actual output spike times $\tilde{t_i}$ and $t_i$, respectively, the learning rule is applied as follows with learning rate $\eta$:
\begin{gather}
\Delta w_{ij} = \eta \left[ \sum_{t_j \in x_j} \lambda(\tilde{t_i} - t_j) - \sum_{t_j \in x_j} \lambda(t_i - t_j) \right].
\label{eq:dw}
\end{gather}
This learning rule takes the difference between two spike-timing coincidence windows, driving a neuron to fire a single spike at its target timing via synaptic weight modification. Specifically, the first term on the right-hand side measures the ability of each input spike at $t_j$ to induce an output spike at $\tilde{t_i}$, while the second term measures the responsibility of each input spike at $t_j$ for causing an output spike at $t_i$. Using this information, the rule changes weights in such a way as to suppress any inputs which contribute to an output spike occurring far from $\tilde{t_i}$, and to reinforce any inputs which will contribute to the emission of an output spike near $\tilde{t_i}$. The rule also works to produce graded weight adjustments when $t_i$ is close to $\tilde{t}_i$, resulting in smooth convergence towards a solution by gradually changing the moment at which the neuron's membrane potential crosses its firing threshold in response to input-evoked \acp{PSP}. In this work only the first output spike from each neuron is used for the learning window, although the \ac{FILT} rule is capable of handling multiple spikes per neuron. In its original formulation the \ac{FILT} rule was devised to train single-layer \acp{SNN}, and tested on randomly-generated input data. 

\subsection{Approximate Backpropagation using \ac{FILT}}
This work uses an augmented version of the \ac{FILT} rule in order to tackle the MNIST classification problem using a multilayer \ac{SNN}. In \cite{Gardner2016} the goal was restricted to training single-layer \acp{SNN} to learn associations between arbitrary input and output spike patterns. The goal in this work is to correctly classify input images of handwritten digits via a first-to-spike decision process in the output layer of a network, as facilitated using precise spike time learning in hidden and output layers. With the ultimate purpose of inducing the correct output neuron to spike earliest in response to its associated class of input data, we use the target spike time $\tilde{t}^l_j$ of each neuron $j$ in a layer $l$ as a training parameter. Rather than being static as in \cite{Gardner2016}, each $\tilde{t}^l_j$ is selected anew for each neuron in each training iteration according to two metrics. These metrics are: (1) the first actual output spike time $t^l_j$, if present, and (2) a figure of merit associated with each neuron which we term the ``desirability'' $d^l_j$. The desirability reflects a measure, relative to the other neurons in layer $l$, of how helpful an early spike from neuron $j$ will be. The desirability of every neuron in each layer lies within the range $[-1,1]$ and is computed recursively with the output layer $l=L$ as the base case. For classification, the output layer neuron which is assigned to the correct class has a default desirability of 1 while all other outputs have -1. For neurons in any non-output layer, $1 \leq l < L$, the desirability is calculated using the desirabilities of the neurons in layer $l+1$ and the corresponding weights $w^{l+1}_{ij}$. Specifically, the vector of pre-normalized desirabilities for all $N_l$ neurons in layer $l$ is $\boldsymbol{\tilde{d}^l} \in \mathbb{R}^{N_l}$, calculated as
\begin{gather}
    \boldsymbol{\tilde{d}^l} = (\boldsymbol{w}^{l+1})^T \cdot \boldsymbol{d}^{l+1},
\end{gather}
where $\boldsymbol{w}^{l+1} \in \mathbb{R}^{N_{l+1} \times N_l}$ is the matrix of weights connecting layer $l$ to layer $l+1$, and $\boldsymbol{d}^{l+1} \in \mathbb{R}^{N_{l+1}}$ is the vector of desirabilities in layer $l+1$. Bold symbols are used to indicate vector or matrix quantities. The $\boldsymbol{\tilde{d}^l}$ vector is then normalized:
\begin{gather}
    \boldsymbol{d}^l = 1 + 2 \frac{\boldsymbol{\tilde{d}^l} - max(\boldsymbol{\tilde{d}^l})}{max(\boldsymbol{\tilde{d}^l}) - min(\boldsymbol{\tilde{d}^l})}
\end{gather}
This shifts the desirabilities in layer $l$ to fall within the range $[-1, 1]$, preventing signal decay in a network with many hidden layers. The relative rankings of the neurons in layer $l$ are preserved by this normalization. The lowest $\tilde{d}^l_j$ in the layer becomes $d^l_j = -1$ and the highest becomes $d^l_j = +1$. The desirability of a given neuron thus gives a relative measure of how strongly that neuron will excite (suppress) neurons in layer $l + 1$ that themselves have a high (low) desirability.

The desirability rule is somewhat analogous to a typical \ac{SNN} error backpropagation algorithm in that it contains a measure of how beneficial each neuron is given the desired outcome of the final layer. However, unlike true backpropagation the desirability rule does not explicitly compute the error gradient with respect to each weight. Instead a much simpler calculation is used, thereby saving significantly on computing power. This also equips the network to counteract the vanishing gradient problem for deep networks.

After computing $\boldsymbol{d}^l$, the neurons are all trained via the \ac{FILT} rule (\ref{eq:dw}) with
\begin{gather}
\tilde{t}^l_j = 
\begin{cases}
  t^l_j - \delta t  & \text{for } d^l_j \ge d_t\\    
  \infty  & \text{for } d^l_j < d_t,
\end{cases}
\label{eq:dt}
\end{gather}
where $d_t$ is the desirability threshold. If $t^l_j$ does not exist, it is substituted by $5\cdot l$ ms. The purpose of the target shift term $\delta t$ is to draw forward the spike times of the most desirable neurons to distinguish them from the others. In turn, the spike times of the neurons in layer $l + 1$ connected most strongly to the desirable neurons in layer $l$ will be similarly drawn forward. Ultimately the ideal output neuron will be encouraged to spike earlier than its neighbors, thus correctly classifying the input. The impact of the target shift $\delta t$ for simple systems will be examined in Sec. \ref{sec:Simulation}. 

\subsection{Additional Learning Constraints}
To accompany the \ac{FILT} rule we also apply RMSprop \cite{Hinton2012} for synapse-specific, adaptive learning rates, providing a means to modulate the magnitude of weight changes. RMSprop has successfully been applied to multilayer \ac{SNN} training in \cite{Zenke2018}, motivating our choice here. With a typical value of $\beta = 0.9$, we calculate the RMSprop scaling term $R_b$ for each batch $b$ as 
\begin{gather}
\boldsymbol{R_b} = \beta \cdot \boldsymbol{R_{b-1}} + (1-\beta)\cdot \boldsymbol{\Delta w_b}^2,
\label{eq:RMSprop1}
\end{gather}
where $\boldsymbol{\Delta w_b}$ is the set of all basic weight changes computed according to (\ref{eq:dw}) in batch $b$ and $\boldsymbol{R_0} = \boldsymbol{1}$. Then the weights $\boldsymbol{w_b}$ are updated:
\begin{gather}
\boldsymbol{w_b} = \boldsymbol{w_{b-1}} + \eta \cdot \frac{\boldsymbol{\Delta w_b}}{\sqrt{\epsilon + \boldsymbol{R_b}}},
\label{eq:RMSprop2}
\end{gather}
where $\eta = 0.01$ and $\epsilon = 0.003$ are the learning rate and RMSprop offset term, respectively. The offset $\epsilon$ is used to avoid singularities when $\boldsymbol{R_b}$ has entries close to zero. 

Synaptic scaling is also applied to the hidden layers to ensure that a reasonable level of spiking activity is present. Without this the hidden units may stop spiking entirely and thus present no activity to excite subsequent layers. The final weights for each batch $\boldsymbol{w_b}*$ are computed with an additional scaling term:
\begin{gather}
\boldsymbol{w_b}* = \boldsymbol{w_b} + \gamma \cdot |\boldsymbol{w_{b-1}}*|(1 - \boldsymbol{S_b})
\label{eq:synaptic scaling}
\end{gather}
where $\gamma = 0.01$ is the scaling coefficient, $\boldsymbol{S_b}$ is a matrix containing the number of spikes produced by each neuron. The precise matrix manipulations required to match the dimensions of $\boldsymbol{S_b}$ and $\boldsymbol{w_b}$ are omitted for brevity. A dropout rate of 35 \% is also applied to the network during training. Dropout has been shown to be useful in preventing overfitting\cite{Hinton2012Improving}.

\section{Simulation and Results}
\label{sec:Simulation}

\begin{figure}
    \centering
    \includegraphics[scale=0.43]{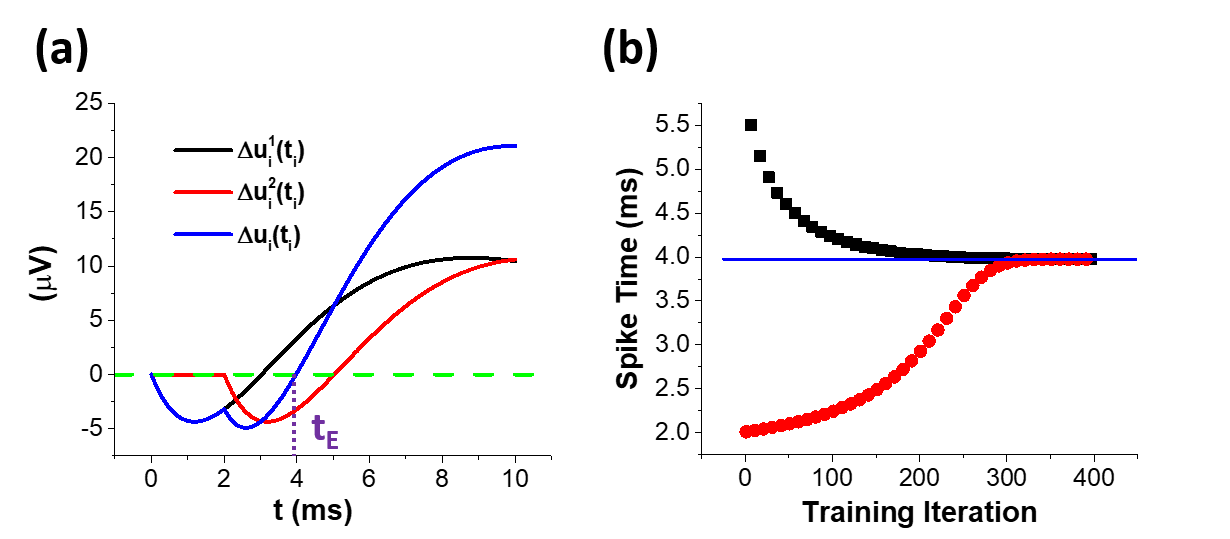}
    \caption{(a) The predicted change in a neuron's membrane potential $u_i$ at its actual firing time $t$ when trained with the \ac{FILT} rule for a single iteration with $\delta t$ = 0.5 ms. The neuron is stimulated by two input spikes occurring at 0 and 2 ms, each of which is received through a different synaptic weight. The terms $\Delta u_i^1$ and $\Delta u_i^2$ indicate the individual contributions to the net change $\Delta u_i$, which result from training on the first and second input spikes, respectively. An equilibrium point exists at $t = t_E$ when $\Delta u_i$ is zero. (b) Two simulated spike rasters showing that the spike time of a single neuron trained on these inputs approaches $t_E$ whether the initial spike time is before or after $t_E$, demonstrating that $t_E$ is an attractor point.}
    \label{fig:dvFig}
\end{figure}

\begin{figure}
    \centering
    \includegraphics[scale=0.43]{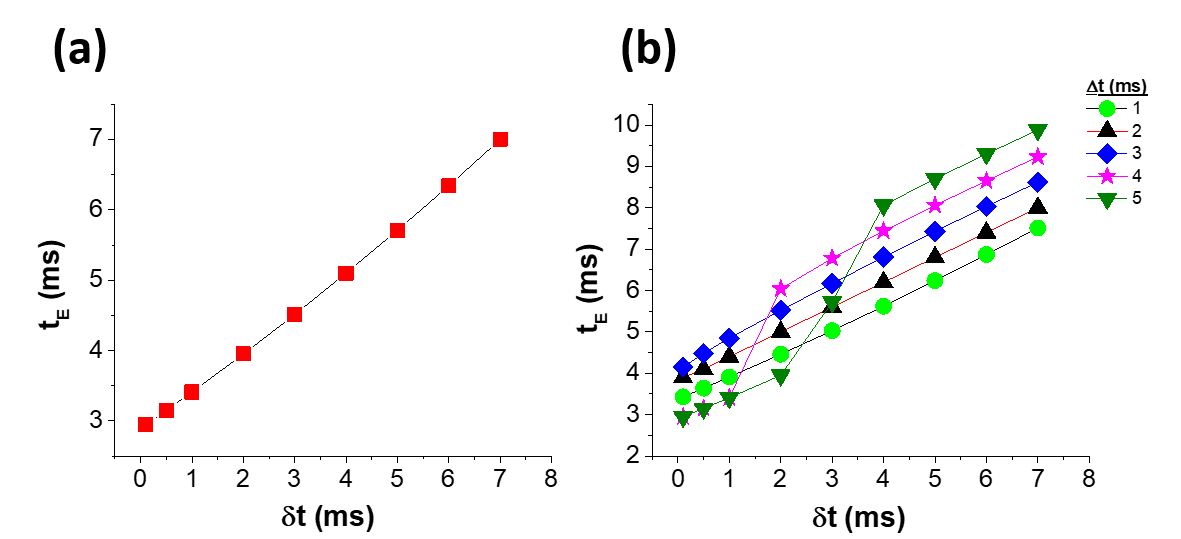}
    \caption{(a) Attractive equilibrium spike time $t_E$ for a single-input neuron as a function of the applied training shift $\delta t$. Counterintuitively, $t_E$ is positively correlated with the $\delta t$. (b) $t_E$ for a two-input neuron with input spikes at $[0, \Delta t]$ as a function of $\delta t$ where $\Delta t$ is a parameter. Provided that $\Delta t$ is not too much larger than $\delta t$ so that there is some training window overlap, the resulting $t_E$ is positively correlated with $\Delta t$. Overall, the greater the delay between inputs and the greater the training shift, the more delayed the equilibrium spike time will be.}
    \label{fig:teFig}
\end{figure}

\begin{figure}
    \centering
    \includegraphics[scale=0.8]{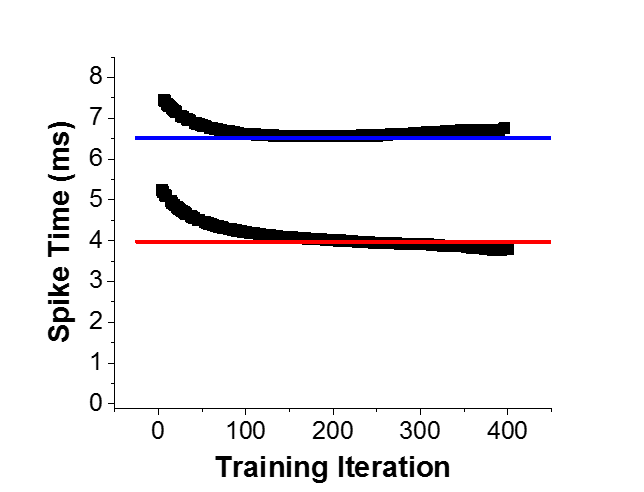}
    \caption{Spike raster for a neuron trained with two different spike trains, one of which is chosen randomly for each training iteration. The solid lines indicate the ideal equilibrium time $t_E$ corresponding to the two spike trains.}
    \label{fig:2trainsFig}
\end{figure}

\begin{figure*}
    \centering
    \includegraphics[scale=0.7]{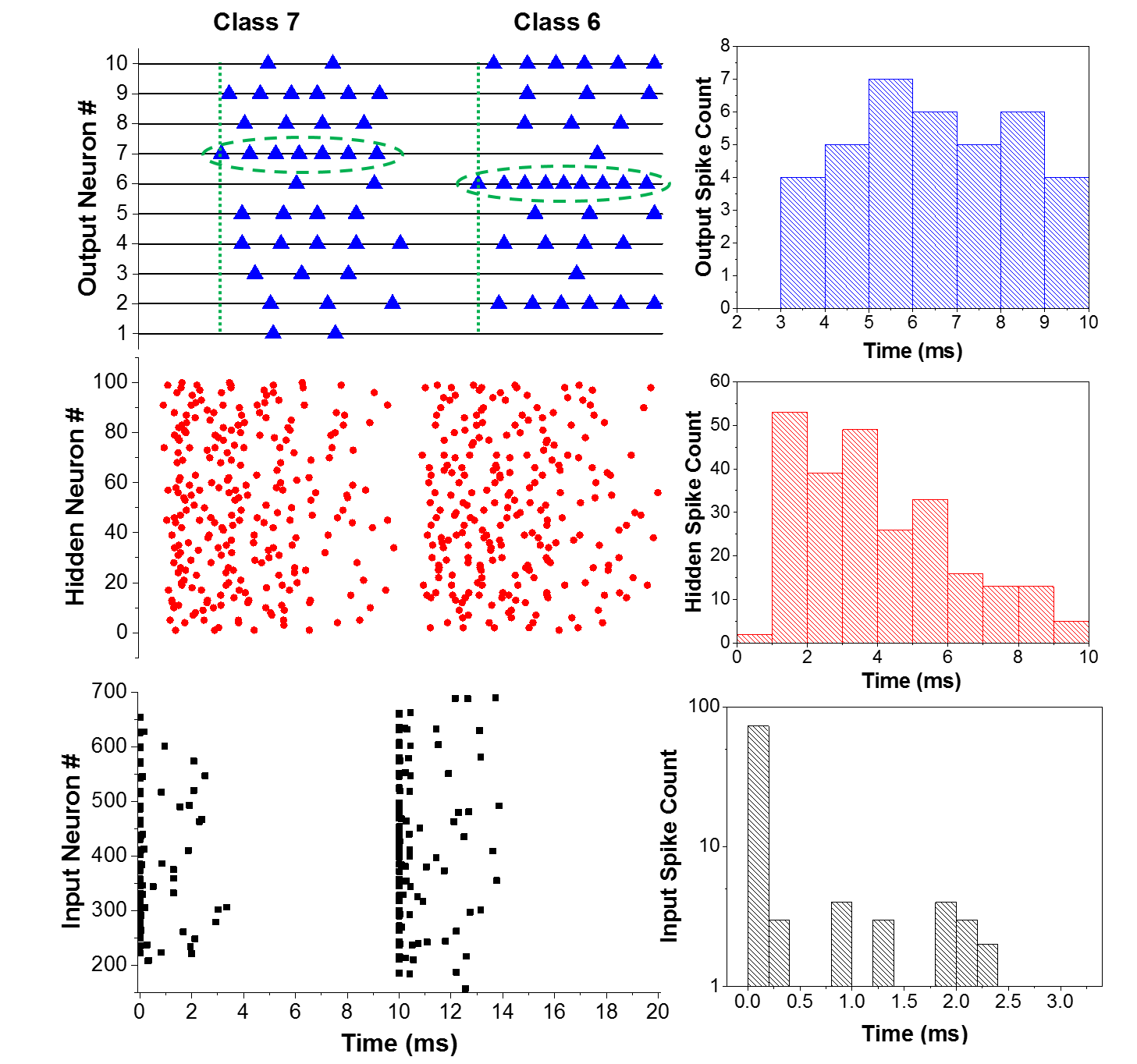}
    \caption{Spike rasters for the input, hidden and output layers arising from two input images presented at 0 and 10 ms. A histogram of the spike timings for each layer is included as well.}
    \label{fig:rasterFig}
\end{figure*}

\subsection{Single Neuron Response}
We first study the behavior of a simple one-neuron network as described in Sec. \ref{sec:Network Model} B. The neuron is presented with one or more arbitrary input spike trains injected through two channels, and its weights updated according to the \ac{FILT} rule as defined by (\ref{eq:dw}) with a target shift term $\delta t$ as in (\ref{eq:dt}). Suppose each input channel contributes a single spike, with timings $t_1$ and $t_2$ for the first and second channels, respectively. The membrane potential of the neuron, excluding the reset kernel, is then described by
\begin{gather}
    u_i(t) = w_1\cdot \epsilon(t-t_1) + w_2\cdot \epsilon(t - t_2).
\end{gather}
If the neuron spikes at $t_i$ and its weights subsequently updated according to the \ac{FILT} rule, with its target firing time shifted forwards by $\delta t$ with respect to $t_i$, then its membrane potential at $t_i$ changes by
\begin{gather*} 
\Delta u_i(t_i) = \Big(\lambda(t_i - \delta t - t_1) - \lambda(t_i - t_1)\Big)\cdot \epsilon(t_i - t_1) \text{ }+ 
\end{gather*}
\begin{gather}
    \Big(\lambda(t_i - \delta t - t_2) - \lambda(t_i - t_2)\Big)\cdot \epsilon(t_i - t_2).
\end{gather}
In Fig. \ref{fig:dvFig}(a) we show $\Delta u_i(t_i)$ as a function of $t_i$ for a sample selection of spike times: $t_1 = \SI{0}{ms}$ and $t_2 = \SI{2}{ms}$, and $\delta t = \SI{0.5}{ms}$. This includes the individual contributions from the two input channels. There is a clear equilibrium point $t_E$ at which $\Delta u_i(t_i) = 0$, indicating that if $t_i = t_E$ the membrane potential of the neuron will not change as a result of applying \ac{FILT}. However, if $t_i > t_E$ then $\Delta u_i(t_i)$ will be positive, causing $t_i$ to decrease in the next iteration. Alternatively, if $t_i < t_E$ then $\delta u_i(t_i)$ becomes negative, leading to an increase in $t_i$. Thus we conclude that $t_E$ is an attractive equilibrium for $t_i$ under the \ac{FILT} rule with positive $\delta t$. To test this conclusion we simulated several trials involving a single neuron with two input channels contributing spikes at $[0, 2] \mathrm{ms}$ and randomly initialized weights. Fig. \ref{fig:dvFig}(b) shows the spike rasters for the neurons in two such trials where the initial spike was either earlier or later than $t_E$. In order to further understand the dynamics of a neuron being trained with a target shift, we calculated the predicted $t_E$ value for various $\delta t$, for both a single-input neuron and a two-input neuron with the second input spike occurring an arbitrary time $\Delta t$ after the first input spike (see Fig. \ref{fig:teFig}). We note that, counterintuitively, the larger the applied $\delta t$ the greater $t_E$ becomes. Adding a second input spike has the effect of shifting the $t_E$ curve upwards when $\delta t$ is sufficiently large that the corresponding single-input $t_E$ value occurs after the second input spike. Prior to this point, the second input has no influence. In Fig. \ref{fig:2trainsFig} we show a neuron learning with two different input spike trains. Each input train has a unique $t_E$ to which the output spike time would converge if trained solely on that input. For each training iteration, one of the two spike trains is randomly selected and presented to the network as an input. Despite having to match two different input sets, the neuron does remarkably well in converging toward each $t_E$ in response to each input train, as indicated by the solid lines in the figure.

We believe this discussion serves to illustrate the potential usefulness of setting the target time $\tilde{t}_i = t_i - \delta t$ term within the \ac{FILT} rule. In many supervised spiking neuron training algorithms, a specific, absolute target time must be chosen. In any reasonable neuron model there will be some delay between a presynaptic spike and the corresponding \ac{PSP} peak. With multiple inputs and multiple layers, it can therefore be difficult to judge the optimal time for each neuron to spike in response to its particular inputs even prior to considering proper network output. The target shift addition to the \ac{FILT} rule allows each neuron to find its own optimal spike time in response to each different input pattern, similar to an unsupervised neuron. However we are also able, in a reinforcement manner, to restrict the spiking of specific neurons by modifying their target time according to the desirability value as discussed above.

\begin{figure}
    \centering
    \includegraphics[scale=0.45]{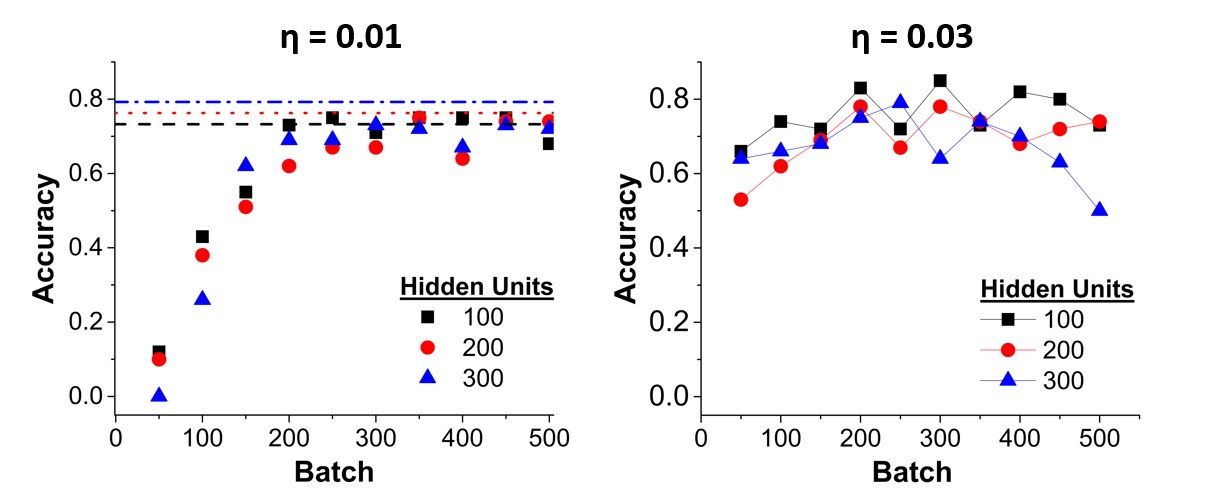}
    \caption{Classification accuracy of the \acp{SNN} with the number of hidden units as a parameter. For comparison the dashed lines, from top to bottom, represent the final test accuracy of the Tensorflow networks with 300, 200 and 100 hidden units respectively. Two different values of the learning rate $\eta$ were tested. The network with 100 hidden units appears to have the best performance.}
    \label{fig:N1Fig}
\end{figure}

\begin{figure}
    \centering
    \includegraphics[scale=0.45]{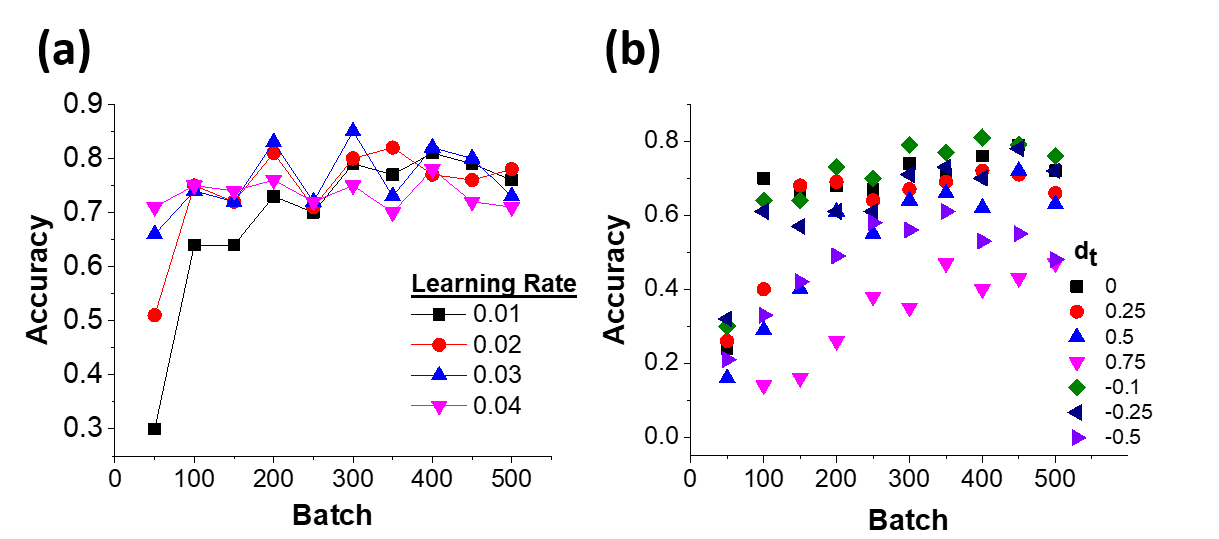}
    \caption{(a) Classification accuracy of the \acp{SNN} with 100 hidden units, $d_t$ = -0.1 and varying $\eta$. The best performance is achieved with $0.03$ although the difference is slight. (b) Classification accuracy with $\eta = 0.01$ and $d_t$ as a parameter. The best performance results from $d_t = -0.1$.}
    \label{fig:dtFig}
\end{figure}

\subsection{MNIST Classification}
We now study the classification of images containing handwritten digits provided by MNIST. An example of the MNIST classification network output is shown in Fig. \ref{fig:rasterFig}. The images of a `7' and a `6' are provided to the network at 0 and 10 ms respectively. The images are transformed into input spikes as described in Section II C., which mostly cluster near 0 and 10 ms. The hidden neurons encode this information more evenly across the timing windows. Despite learning rules which attempt to enforce one spike per output neuron, some images such as the ones tested in Fig. \ref{fig:rasterFig} result in multiple output spikes. Despite this the two samples provided to the network are both correctly classified. The 7th and 6th neurons, respectively, are the first to spike, as well as being the most active during their respective windows. We tested networks with multiple different hidden layer sizes, learning rates and $d_t$ thresholds. Each network was trained for 500 batches with batch sizes of either 20 or 100 samples each. There was no significant difference in results noted between the two batch sizes. For comparison, a standard fully-connected non-spiking network of equivalent size was trained in TensorFlow using sigmoidal activation neurons which we treat as roughly equivalent to spiking neurons. This Tensorflow network utilized full error-backpropagation training. The mean test accuracy using ensembles of 10 networks with 100, 200 and 300 hidden units respectively is plotted against the batch number in Fig. \ref{fig:N1Fig} alongside dashed lines that indicate the final test accuracy of the TensorFlow networks. In these simulations $d_t = 0$. The three trials very nearly approach the accuracy of the backpropagation networks. For 100 hidden units the optimal learning rate is shown to be 0.03 as indicated in Fig. \ref{fig:dtFig}(a) although the variation between $\eta = 0.01$ and $\eta = 0.04$ is slight. In additional trials that explored $\eta$ values outside this range the performance dropped off sharply. In Fig. \ref{fig:dtFig}(b) we study the effect of varying $d_t$ for the 100-unit network. The best case appears to be a $d_t = -0.1$, indicating that roughly the top 45th percentile of hidden neurons should be reinforced and the rest suppressed. Indeed the best trial matches or outperforms even the larger 300-unit backpropagation network shown in Fig. \ref{fig:N1Fig}. 

\section{Discussion}
\label{sec:Discussion}

We have devised and demonstrated a new supervised learning rule for classifying images with spiking neural networks. We would like to begin the discussion by highlighting and reiterating several of its important aspects. 

First, this rule begins with \ac{FILT}, a single-layer supervised learning algorithm, and modulates it using the backpropagated desirability as a fitness signal. Imprecise backpropagation, relying on approximated feedback signalling, has previously been shown to be an effective learning method\cite{Gardner_2015,lillicrap2016random}. The \ac{FILT} rule, among some others like it, is derived using a rate-substitution method which solves the discontinuous spike-gradient problem. However unlike some other supervised spike-based learning rules such as \cite{Sporea2013}, \ac{FILT} preserves nonlinearity by assuming nonlinear spiking rate functions. In general, any suitable approximate backpropagated signal could be used in tandem with any functional single-layer learning rule for spiking neurons, leading to a new class of hybrid multilayer learning methods. This class possesses some of the advantages of both full \ac{SNN} backpropagation methods and local learning methods.

Second, this hybrid class has significantly reduced computational complexity due to the lack of a true error function and accompanying gradient calculation. By using only implicit reference to network output, we produce a learning rule that would be especially helpful for deep networks and neuromorphic hardware.

Third, by beginning with a single-layer training rule that seeks to impose a specific output spike train on the neurons and using $\tilde{t}^l_j = t^l_j - \delta_t$ we allow the chosen neurons to find their own natural spike time. This maintains some of the degrees of freedom in the network while still imposing a desired pattern by selecting which neurons are trained according to a finite $\tilde{t}^l_j$ via the $d^l_j$ and $d_t$ values. This bears some relation to \ac{STDP}, a timing-based rule that many \ac{SNN} implementations in hardware employ\cite{sengupta17,kim15,Furber2014,Friedmann2017,Lin2018}. Hardware built to natively implement \ac{STDP} should therefore be more easily adapted to use the reinforced, timing-based \ac{FILT} rule as opposed to a full backpropagation method. An example is Heidelberg's \ac{DLS} platform, which not only implements local, \ac{STDP}-like, pre- and postsynaptic correlation traces, but also includes an embedded microprocessor dedicated to plasticity processing that is capable of combining local traces with a third factor, such as an external reward signal \cite{Friedmann2017}. \ac{DLS} also operates at a large speed-up factor of 1000 compared to the biological time-scale it emulates, making it ideally suited to long-term \ac{SNN} simulations.

Interestingly, we observe that our hybrid learning approach shares certain similarities with a reward-maximization procedure as studied in \cite{Xie2004,Pfister2006,Florian2007,Fremaux2010}. In more detail, we first note that the \ac{FILT} rule approximates the supervised, maximum-likelihood approach to learning for a probabilistic neuron model, but taken in the limit of a deterministic system \cite{Gardner2016}. We then consider that, as described in \cite{Pfister2006,Fremaux2010}, if the target spike train of a neuron trained by maximum-likelihood is substituted with its actual output spike train, then the learning rule becomes an unsupervised one. Hence, if this unsupervised rule is combined with an external `success signal' to guide weight changes, then the rule works in such a way as to maximize the likelihood of a neuron generating a spike train which positively correlates with the receipt of a positive-valued success signal: a process that is otherwise referred to as reward-maximization \cite{Fremaux2010}. Taking these points into account, we relate our hybrid learning method based on \ac{FILT} to the reward-maximization method for the following reasons:
\begin{itemize}
    \item In our implementation the target firing time of a neuron is not explicitly supervised: the prescribed target timing actually depends on the actual output firing time of the neuron, but shifted earlier by a small amount.
    \item When unsupervised the rule works to progressively shift the actual firing time of a neuron earlier, but since the \ac{FILT} rule is additionally modulated by an external `desirability' signal that is linked to the overall success of the network, the firing time is instead adjusted to shift either forwards or backwards according to how this influences the network's goal.
    \item Since the goal of the network is to drive an early response in one of the output layer neurons according to its associated class label, then the communicated desirability works to shift the strongest contributing hidden spikes earlier.
    \item Although the evaluation of the modulatory signals used here involves more computational steps than those considered in \cite{Fremaux2010}, they essentially share the same end result of providing task-specific feedback in order to guide desirable weight changes.
\end{itemize}
Hence, it is for the above reasons that we can interpret our hybrid training method as driving a form of reward-maximization during network training, by a combination of layer-wise local learning factors modulated by external signalling. It is also notable that reward-modulation of local synaptic activity factors is considered a biologically-plausible hypothesis for learning \cite{Vasilaki2009}, where for example the neuromodulator dopamine is hypothesized to encode a reward-prediction error signal in the brain \cite{Schultz2000}. 

In conclusion, we note that although our experiments demonstrated reasonably high accuracy of our hybrid training method on MNIST, this did not reach the levels of some other, more refined spiking classifier implementations such as in \cite{Connor2013,Mostafa2017}. However, we believe this sufficiently demonstrates proof of concept for the hybrid learning rule we pioneer herein. The results achieved here are a lower bound on the possible performance of the rule, as different choices of reinforcement signal and layer-wise learning rule and additional tuning of the parameters can further enhance the network accuracy.

\bibliographystyle{IEEEtran}
\bibliography{refs}

\begin{thebibliography}{10}
\providecommand{\url}[1]{#1}
\csname url@samestyle\endcsname
\providecommand{\newblock}{\relax}
\providecommand{\bibinfo}[2]{#2}
\providecommand{\BIBentrySTDinterwordspacing}{\spaceskip=0pt\relax}
\providecommand{\BIBentryALTinterwordstretchfactor}{4}
\providecommand{\BIBentryALTinterwordspacing}{\spaceskip=\fontdimen2\font plus
\BIBentryALTinterwordstretchfactor\fontdimen3\font minus
  \fontdimen4\font\relax}
\providecommand{\BIBforeignlanguage}[2]{{%
\expandafter\ifx\csname l@#1\endcsname\relax
\typeout{** WARNING: IEEEtran.bst: No hyphenation pattern has been}%
\typeout{** loaded for the language `#1'. Using the pattern for}%
\typeout{** the default language instead.}%
\else
\language=\csname l@#1\endcsname
\fi
#2}}
\providecommand{\BIBdecl}{\relax}
\BIBdecl

\bibitem{Maass1997}
W.~Maass, ``Networks of spiking neurons: the third generation of neural network
  models,'' \emph{Neural Networks}, vol.~10, no.~9, pp. 1659--1671, 1997.

\bibitem{Bohte2002}
S.~M. Bohte, J.~N. Kok, and H.~La~Poutre, ``Error-backpropagation in temporally
  encoded networks of spiking neurons,'' \emph{Neurocomputing}, vol.~48, no.
  1-4, pp. 17--37, 2002.

\bibitem{Sporea2013}
I.~Sporea and A.~Gr{\"u}ning, ``Supervised learning in multilayer spiking
  neural networks,'' \emph{Neural Computation}, vol.~25, no.~2, pp. 473--509,
  2013.

\bibitem{Ponulak2010}
F.~Ponulak and A.~Kasi{\'n}ski, ``Supervised learning in spiking neural
  networks with resume: sequence learning, classification, and spike
  shifting,'' \emph{Neural Computation}, vol.~22, no.~2, pp. 467--510, 2010.

\bibitem{Gardner2015}
B.~Gardner, I.~Sporea, and A.~Gr{\"u}ning, ``Learning spatiotemporally encoded
  pattern transformations in structured spiking neural networks,'' \emph{Neural
  Computation}, vol.~27, no.~12, pp. 2548--2586, 2015.

\bibitem{Pfister2006}
J.-P. Pfister, T.~Toyoizumi, D.~Barber, and W.~Gerstner, ``Optimal
  spike-timing-dependent plasticity for precise action potential firing in
  supervised learning,'' \emph{Neural Computation}, vol.~18, no.~6, pp.
  1318--1348, 2006.

\bibitem{Mostafa2017}
H.~Mostafa, ``Supervised learning based on temporal coding in spiking neural
  networks,'' \emph{IEEE Transactions on Neural Networks and Learning Systems},
  vol.~29, no.~7, pp. 3227--3235, 2017.

\bibitem{Zenke2018}
F.~Zenke and S.~Ganguli, ``Superspike: Supervised learning in multilayer
  spiking neural networks,'' \emph{Neural Computation}, vol.~30, no.~6, pp.
  1514--1541, 2018.

\bibitem{vanRossum2001}
M.~C.~W. van Rossum, ``A novel spike distance,'' \emph{Neural Computation},
  vol.~13, no.~4, pp. 751--763, 2001.

\bibitem{Gerstner2002}
W.~Gerstner and W.~M. Kistler, \emph{Spiking neuron models: Single neurons,
  populations, plasticity}.\hskip 1em plus 0.5em minus 0.4em\relax Cambridge
  University Press, 2002.

\bibitem{Florian2012}
R.~V. Florian, ``The chronotron: a neuron that learns to fire temporally
  precise spike patterns,'' \emph{PLoS ONE}, vol.~7, no.~8, p. e40233, 2012.

\bibitem{Mohemmed2012}
A.~Mohemmed, S.~Schliebs, S.~Matsuda, and N.~Kasabov, ``Span: Spike pattern
  association neuron for learning spatio-temporal spike patterns,''
  \emph{International Journal of Neural Systems}, vol.~22, no.~04, p. 1250012,
  2012.

\bibitem{Memmesheimer2014}
R.-M. Memmesheimer, R.~Rubin, B.~P. {\"O}lveczky, and H.~Sompolinsky,
  ``Learning precisely timed spikes,'' \emph{Neuron}, vol.~82, no.~4, pp.
  925--938, 2014.

\bibitem{Gardner2016}
B.~Gardner and A.~Gr{\"u}ning, ``Supervised learning in spiking neural networks
  for precise temporal encoding,'' \emph{PLoS ONE}, vol.~11, no.~8, p.
  e0161335, 2016.

\bibitem{lecun1998}
Y.~LeCun, L.~Bottou, Y.~Bengio, P.~Haffner \emph{et~al.}, ``Gradient-based
  learning applied to document recognition,'' \emph{Proceedings of the IEEE},
  vol.~86, no.~11, pp. 2278--2324, 1998.

\bibitem{Bohte2002unsupervised}
S.~M. Bohte, H.~La~Poutr{\'e}, and J.~N. Kok, ``Unsupervised clustering with
  spiking neurons by sparse temporal coding and multilayer rbf networks,''
  \emph{IEEE Transactions on Neural Networks}, vol.~13, no.~2, pp. 426--435,
  2002.

\bibitem{Hung2005}
C.~P. Hung, G.~Kreiman, T.~Poggio, and J.~J. DiCarlo, ``Fast readout of object
  identity from macaque inferior temporal cortex,'' \emph{Science}, vol. 310,
  no. 5749, pp. 863--866, 2005.

\bibitem{Hinton2012}
G.~Hinton, N.~Srivastava, and K.~Swersky, ``Neural networks for machine
  learning lecture 6a overview of mini-batch gradient descent,''
  \emph{Coursera}, vol.~14, p.~8, 2012.

\bibitem{Hinton2012Improving}
G.~E. Hinton, N.~Srivastava, A.~Krizhevsky, I.~Sutskever, and R.~R.
  Salakhutdinov, ``Improving neural networks by preventing co-adaptation of
  feature detectors,'' \emph{arXiv preprint arXiv:1207.0580}, 2012.

\bibitem{Gardner_2015}
B.~Gardner, I.~Sporea, and A.~Grüning, ``Learning spatiotemporally encoded
  pattern transformations in structured spiking neural networks,'' \emph{Neural
  Computation}, vol.~27, no.~12, p. 2548–2586, Dec 2015.

\bibitem{lillicrap2016random}
T.~P. Lillicrap, D.~Cownden, D.~B. Tweed, and C.~J. Akerman, ``Random synaptic
  feedback weights support error backpropagation for deep learning,''
  \emph{Nature communications}, vol.~7, p. 13276, 2016.

\bibitem{sengupta17}
A.~Sengupta and K.~Roy, ``Encoding neural and synaptic functionalities in
  electron spin: A pathway to efficient neuromorphic computing,'' \emph{Applied
  Physics Reviews}, vol.~4, no.~4, p. 041105, 2017.

\bibitem{kim15}
Y.~Kim, Y.~Zhang, and P.~Li, ``A reconfigurable digital neuromorphic processor
  with memristive synaptic crossbar for cognitive computing,'' \emph{ACM
  Journal on Emerging Technologies in Computing Systems (JETC)}, vol.~11,
  no.~4, p.~38, 2015.

\bibitem{Furber2014}
S.~B. Furber, F.~Galluppi, S.~Temple, and L.~A. Plana, ``The spinnaker
  project,'' \emph{Proceedings of the IEEE}, vol. 102, no.~5, pp. 652--665,
  2014.

\bibitem{Friedmann2017}
S.~Friedmann, J.~Schemmel, A.~Gr{\"u}bl, A.~Hartel, M.~Hock, and K.~Meier,
  ``Demonstrating hybrid learning in a flexible neuromorphic hardware system,''
  \emph{IEEE Transactions on Biomedical Circuits and Systems}, vol.~11, no.~1,
  pp. 128--142, 2017.

\bibitem{Lin2018}
C.-K. Lin, A.~Wild, G.~N. Chinya, Y.~Cao, M.~Davies, D.~M. Lavery, and H.~Wang,
  ``Programming spiking neural networks on intel’s loihi,'' \emph{Computer},
  vol.~51, no.~3, pp. 52--61, 2018.

\bibitem{Xie2004}
X.~Xie and H.~S. Seung, ``Learning in neural networks by reinforcement of
  irregular spiking,'' \emph{Physical Review E}, vol.~69, no.~4, p. 041909,
  2004.

\bibitem{Florian2007}
R.~V. Florian, ``Reinforcement learning through modulation of
  spike-timing-dependent synaptic plasticity,'' \emph{Neural Computation},
  vol.~19, no.~6, pp. 1468--1502, 2007.

\bibitem{Fremaux2010}
N.~Fr{\'e}maux, H.~Sprekeler, and W.~Gerstner, ``Functional requirements for
  reward-modulated spike-timing-dependent plasticity,'' \emph{Journal of
  Neuroscience}, vol.~30, no.~40, pp. 13\,326--13\,337, 2010.

\bibitem{Vasilaki2009}
E.~Vasilaki, N.~Fr{\'e}maux, R.~Urbanczik, W.~Senn, and W.~Gerstner,
  ``Spike-based reinforcement learning in continuous state and action space:
  when policy gradient methods fail,'' \emph{PLoS Computational Biology},
  vol.~5, no.~12, p. e1000586, 2009.

\bibitem{Schultz2000}
W.~Schultz, ``Multiple reward signals in the brain,'' \emph{Nature Reviews
  Neuroscience}, vol.~1, no.~3, p. 199, 2000.

\bibitem{Connor2013}
P.~O'Connor, D.~Neil, S.-C. Liu, T.~Delbruck, and M.~Pfeiffer, ``Real-time
  classification and sensor fusion with a spiking deep belief network,''
  \emph{Frontiers in Neuroscience}, vol.~7, p. 178, 2013.

\end{thebibliography}


\end{document}